\documentclass[conference]{IEEEtran}

\usepackage[utf8]{inputenc}
\usepackage[square,sort,comma,numbers]{natbib}
\usepackage{float} 
\usepackage{subfig}
\usepackage{caption}
\usepackage{listings}
\usepackage{booktabs,threeparttable}
\usepackage[table,xcdraw]{xcolor}
\usepackage{multirow}
\usepackage{verbatimbox}
\usepackage[nolist]{acronym}
\newacro{AI}{Artificial Intelligence}
\newacro{AIMDEO}{AI Model and Dataset Exchange Ontology}
\newacro{AIMDEP}{AI Model and Dataset Exchange Platform}
\newacro{API}{Application Programming Interface}
\newacro{CSV}{Comma-separated Values}
\newacro{EDA}{Electronic Design Automation}
\newacro{EMMM}{Experiment Management Meta-Model}
\newacro{ITO}{Intelligence Task Ontology and Knowledge Graph}
\newacro{IP}{Intellectual Property}
\newacro{JSON}{JavaScript Object Notation}
\newacro{LD}{Linked Data}
\newacro{OWL}{Web Ontology Language}
\newacro{RDF}{Resource Description Framework}
\newacro{REST}{Representational State Transfer}
\newacro{URI}{Uniform Resource Identifier}
\newacro{UI}{User Interface}
\newacro{MVT}{Model-View-Template}
\usepackage{balance}
\usepackage{eso-pic} 
\usepackage[hidelinks]{hyperref} 
\usepackage{tikz}
\newcommand*\circled[1]{\tikz[baseline=(char.base)]{
            \node[shape=circle,draw,inner sep=1pt] (char) {#1};}}

\lstdefinelanguage{Turtle}{
    sensitive=false,
    morestring=[s]{"}{"},
    stringstyle=\color{black}\bfseries,
    showstringspaces=false
}

\definecolor{backcolour}{rgb}{0.96,0.96,0.96}

\newcommand{\IEEEAcceptedManuscriptNotice}{%
  \AddToShipoutPictureFG*{%
    \AtPageLowerLeft{%
      \put(48,9){%
        \parbox[b]{516pt}{%
          \hrule
          \vspace{2pt}
          \fontsize{5.0}{5.5}\selectfont
          \raggedright
          \textcopyright~2024 IEEE. Personal use of this material is permitted.
          Permission from IEEE must be obtained for all other uses, in any current
          or future media, including reprinting/republishing this material for
          advertising or promotional purposes, creating new collective works, for
          resale or redistribution to servers or lists, or reuse of any copyrighted
          component of this work in other works.

          \textit{This is the accepted manuscript of: J. Novacek, A. Ahari,
          T. M{\"u}ller, S. Reiter, A. Viehl, and O. Bringmann,
          ``Ontology-Supported AI Model and Dataset Management,'' in 2024 IEEE
          22nd International Conference on Industrial Informatics (INDIN),
          pp. 1--6, 2024. The version of record is available at
          \href{https://doi.org/10.1109/INDIN58382.2024.10774524}{%
          \textcolor{blue}{https://doi.org/10.1109/INDIN58382.2024.10774524}}.}%
        }%
      }%
    }%
  }%
}
\begin{document}
\IEEEAcceptedManuscriptNotice

\title{Ontology-supported\\AI Model and Dataset Management}

 \author{\IEEEauthorblockN{Jan Novacek\IEEEauthorrefmark{1}, Ali Ahari\IEEEauthorrefmark{1}\IEEEauthorrefmark{2}, Tobias Müller\IEEEauthorrefmark{1}\IEEEauthorrefmark{2}, Sebastian Reiter\IEEEauthorrefmark{1},\\
 Alexander Viehl\IEEEauthorrefmark{1}, Oliver Bringmann\IEEEauthorrefmark{1}\IEEEauthorrefmark{2}}
 \IEEEauthorblockA{\IEEEauthorrefmark{1}FZI Research Center for Information Technology\\
 Haid-und-Neu-Str. 10-14, 76131 Karlsruhe}
 \IEEEauthorblockA{\IEEEauthorrefmark{2}University of Tübingen\\
 Sand 14, 72076 Tübingen}}

\maketitle

%
\begin{abstract}
Recently, there has been a great deal of research into improving \acs{AI} methods and their application.
The main focus is on tracking progress, enabling transparent comparisons, and fostering a more profound understanding of \acs{AI}.
In that process, different organizations generate and use plenty of assets that need to be tracked, traced and managed. Moreover, it is important to discover assets relevant for the task at hand.
This paper presents research aiming to contribute to answering the question of what is required to exchange and manage \acs{AI} models and related assets effectively without semantic gaps in an industrial context.
We introduce a platform for \acs{AI} model exchange, which facilitates the usage, exchange, and analysis of AI models and datasets. The platform incorporates an ontology that can foster a more profound common understanding of what is required in these tasks and help tackle the issues mentioned above.
Finally, we elucidate the utility of the platform through the illustration of a use case in the context of real-time critical systems.

\end{abstract}


%
\section{Introduction}
The widespread application and sheer number of existing \ac{AI} models demand discovering the right assets and building trust in the end-user using the models.
By the time of this writing, the data collection and development of \ac{AI} models is often separated from the final users of the models.
However, distributed and shared use can only be achieved through a clear understanding of the functional scope of \ac{AI} models.
This requires a characterization of \ac{AI} models, which includes basic descriptions of the models and basic information about the data, quality, and context.

In the automotive industry, for instance, \ac{AI} models and related assets are transferred between different tiers along the supply chain.
A predominant issue in this regard is a lack of structured metadata of \ac{AI} models with explicit semantics and a common understanding of such a metadata format.
Popular \ac{AI} communities, like \emph{Hugging Face}\footnote{https://huggingface.co}, for example, host numerous \ac{AI} models and datasets. Information about these resources is provided by semi-structured descriptions, mainly consisting of natural language text of differing kinds of content and amounts.
While some but not all models have structured metadata, the lack of them hinders the shared use and development of \ac{AI} models.
To tackle these issues, this work aims to answer the following research question:
\begin{quote}
    What is required to develop and use an \ac{AI} model collaboratively in an industrial context?
\end{quote}

\begin{figure}
    \centering
    \includegraphics[width=\columnwidth]{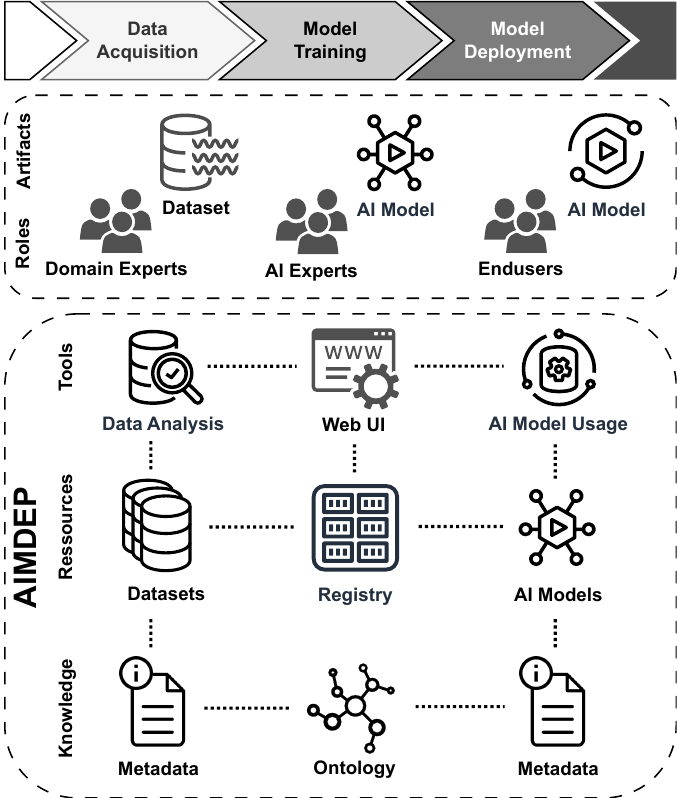}
    \caption{Platform tools, resources, and knowledge for different machine learning development phases}
    \label{fig:aimdep-overview}
\end{figure}

Current state-of-the-art machine learning platforms such as \emph{MLflow}\footnote{https://mlflow.org}, \emph{H2O}\footnote{https://h2o.ai} and \emph{Ray}\footnote{https://www.ray.io} have means to add metadata to models. However, the metadata specifications lack interoperability. 
This problem becomes even more apparent when models need to be exchanged among different companies due to differing metadata content or formats.

In this paper, we present the \emph{\ac{AIMDEP}}, a platform for collaborative \ac{AI} model development and use, see Fig. \ref{fig:aimdep-overview}. This platform incorporates the \emph{\ac{AIMDEO}}, an ontology for the description of machine learning assets, focusing on the expression of metadata of \ac{AI} models and datasets.
The practicability thereof is further demonstrated by a use case.

The main contributions of this paper are:
\begin{itemize}
    \item A platform for collaborative \ac{AI} development and use
    \item An ontology for the description of \ac{AI} models and datasets
    \item Use case showing the applicability of the approach
\end{itemize}


This paper is structured as follows: Section \ref{section:related_work} describes identified related work regarding \ac{AI} asset management as well as \ac{AI} asset characterization.
Section \ref{section:simml} describes the \emph{\ac{AIMDEP}}, followed by a description of the ontology in Section \ref{section:ontology}.
In Section \ref{section:use_cases}, we evaluate the approach with a use case that shows the applicability and benefits of this approach regarding \ac{AI} model and dataset usage.
Finally, Section \ref{section:conclusion} concludes the paper.
%
\section{Related Work}
\label{section:related_work}
This section describes related work and points out differences and similarities to the presented approach. Besides \ac{AI} asset management, \ac{AI} asset characterization is covered.

\subsection{\ac{AI} Asset Management}
The authors in~\cite{idowu2022asset} assessed various tools and platforms for machine learning asset management.
Of all the options, only MLFlow supports asset registry and exchange without requiring data sharing with third parties.
However, it does not support ontologies for the description of metadata of assets.
\emph{MLEM}\footnote{https://mlem.ai} is an open-source tool that provides a standard interface for deploying machine learning models and offers a model registry.
However, its model registry is very limited and does not collect metadata.
Hugging Face is a hub for machine learning models and datasets and offers an interface for deploying the models.

\subsection{\ac{AI} Asset Characterization}
Recent work from Blagec et al. introduced the \emph{\ac{ITO}} \cite{blagec2022curated} which aims primarily to study scientific research but can also be used to annotate and organize information in the \ac{AI} domain. While the \ac{ITO} has the purpose of describing \ac{AI} tasks, the \ac{AIMDEO} ontology presented in this paper is meant to support the exchange of \ac{AI} models and datasets. Moreover, in contrast to the \ac{ITO}, the \ac{AIMDEO} contains concepts for describing the provenance and kind of \ac{AI} model as well as parameters of \ac{AI} models and datasets. Additionally, the ontology presented here allows specifying model evaluation metrics.
Both, the \ac{ITO} and the \ac{AIMDEO} support the specification of intended tasks and subtasks of \ac{AI} models and datasets.

Idowu et al. presented the \emph{\ac{EMMM}} \cite{idowu2022emmm}, which is a metamodel of asset types and their relationships common to the management of machine learning experiments. Besides modeling central concepts such as \textit{models}, \textit{parameters}, or \textit{dependencies}, this metamodel enables the specification of arbitrary metadata for any asset. In addition to the asset structures and their relationships, the metamodel considers version control structures for machine learning as well as traditional assets.
While the metamodel covers central concepts, element metadata is captured using arbitrary key-value mappings, which lacks a common format and explicit semantics.


%
\section{Exchange Platform}
\label{section:simml}
With the advancement of AI in the recent decade, a vast number of AI assets have been produced within various companies and organizations. However, these assets generally lack any additional metadata and are either not discoverable or inaccessible by others. There is also a discrepancy between the environment in which an asset is created and the environment in which other users will use it, which can lead to various compatibility issues and hinder the usage of assets.
To address the abovementioned issues, we have developed an \textit{\acf{AIMDEP}}, which provides a central registry to make AI models and datasets visible and accessible to all users. The \ac{AIMDEP} uses the \ac{AIMDEO} to specify the assets and captures additional metadata for each asset accordingly to make them more interpretable. Moreover, \ac{AIMDEP} supports the export of \ac{AI} model metadata according to the \ac{AIMDEO} in the form of micro-ontologies. Finally, the online deployment integrated into the AIMDEP provides a standard platform-independent interface to deploy the assets. 

\begin{figure}[H]
    \centering
    \includegraphics[width=0.8\columnwidth]{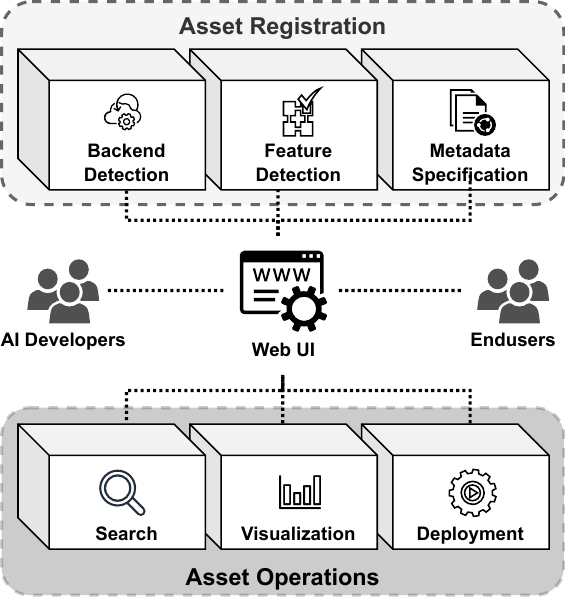}
    \caption{Core features of the platform}
    \label{fig:aimdep-features}
\end{figure}

Fig. \ref{fig:aimdep-features} shows an overview of the core AIMDEP features
regarding asset registration and operations. The platform em-
ploys a client-server architecture, with the server and the and
the Web User Interface (UI) implemented using the Django
framework. Additionally, custom clients can interact with the
server through the provided Representational State Transfer
(REST) Application Programming Interface (API). Leverag-
ing Django’s Model-View-Template (MVT) architecture, any
updates to the ontology in the AIMDEO can seamlessly apply
to the model component of the server without disrupting
its overall functionality. This design ensures the platform’s
flexibility for accommodating future ontology updates.
The communication between the server and the clients is
encrypted, and the access management policy is implemented
on the server side to restrict access to assets to privileged users.
The models and datasets go through a register process to be
added to the AIMDEO-based database. The AIMDEP supports
multiple operations on the registered assets. For both datasets
and models, it offers a semantic search functionality and the
download of assets for offline deployment. The datasets can be
visualized to gain more insight into the data using Plotly [4],
and the models can be deployed online and evaluated using
the interface provided by MLEM [5]. Next, we discuss each
part in more detail.

\subsection{Asset Registration}

The registration of models and datasets starts by uploading the physical asset files to the platform. The AIMDEP attempts to identify the back-end required to access and deploy the asset. This includes identifying the data handlers for the datasets and the framework needed to evaluate and deploy the model. The AIMDEP supports various common dataset formats, including \ac{CSV}, Excel, JSON, and Parquet, and offers support for models exported by the most popular machine learning frameworks, including \emph{Scikit-Learn}\footnote{https://scikit-learn.org/}, \emph{TensorFlow}\footnote{https://www.tensorflow.org}, and \emph{PyTorch}\footnote{https://pytorch.org}. The user can edit the selected framework if required. Any custom dataset and model can also be registered without specifying the framework.
However, the platform would be unable to visualize the dataset or deploy the model later, since the platform requires the framework to interact with the assets.

Next, models and datasets' input and output features are defined semi-automatically. The platform tries to detect the features, but the user should verify and add further metadata for each feature. Each feature is then stored as a \emph{Parameter} based on AIMDEO. The user should also add the configuration parameters of models at this stage. Finally, the user adds the additional metadata following the AIMDEO, which includes, for instance, adding metrics to the model. The asset can then be registered in the central database of the platform for later access and operations.

\subsection{Asset Operations}

The AIMDEP supports multiple asset operations:
For datasets, it offers interactive visualization in the form of statistical tables, various plots, and feature analyses (only for numerical datasets) to identify the importance of each feature. This offers a better insight into the data to train and evaluate efficient models later.
The visualizations are rendered by a \emph{Plotly}\footnote{https://plotly.com} engine on the server side. Since the visualization of the whole dataset can be very time-consuming for massive datasets, the AIMDEP can limit the visualization to a random subset of the dataset with a given size.

Online deployment is offered for the registered models with a supported framework. It enables a simple and platform-independent interface for the inference of the models. The inference of a model can be performed either through the \ac{REST} \ac{API} or through a graphical interface based on \emph{Gradio}\footnote{https://www.gradio.app}, which is generated based on the task and subtask of the model and its input and output features.

The AIMDEP supports the search and download of assets for both models and datasets. The search is powered by \emph{OpenSearch}\footnote{https://opensearch.org} and utilizes the metadata captured for each asset to perform a search and returns the most relevant assets for the query.
Assets can also be downloaded for offline use or integration in custom workflows.
Next to the original files, additional metadata captured for the asset is also provided as micro-ontologies containing the attributes and their values.
Fig. \ref{fig:owl-micro-ontology} shows an example output of an export of \ac{AI} model and corresponding dataset metadata from \ac{AIMDEP} referring to the \ac{AIMDEO}.
%
\section{Ontology}
\label{section:ontology}
The \emph{AI Model and Dataset Exchange Ontology (AIMDEO)} captures essential concepts that are part of collaborative AI model development and use. The ontology is expressed as an \ac{OWL} \cite{antoniou2004webOntologyLanguage} ontology. 
Overall metrics of the ontology are shown in Tab. \ref{table:ontology_metrics}.
\begin{table}[H]
\centering
\begin{tabular}{@{}ll@{}}
\toprule
\textbf{Metric} & \textbf{Amount} \\ \midrule
Axiom count & 414 \\
Logical axioms count & 178 \\
Declaration axioms count & 105 \\
Class count & 47 \\
Object property count & 29 \\
Data property count & 17 \\
Individual count & 5 \\
Annotation property count & 18 \\
DL expressivity & SHOIQ(D) \\ \bottomrule
\end{tabular}
\caption{Ontology metrics}
\label{table:ontology_metrics}
\end{table}

\subsection{Collaboration Scenarios}
\label{section:methodology}
\begin{figure}
  \centering
  \includegraphics[width=\linewidth]{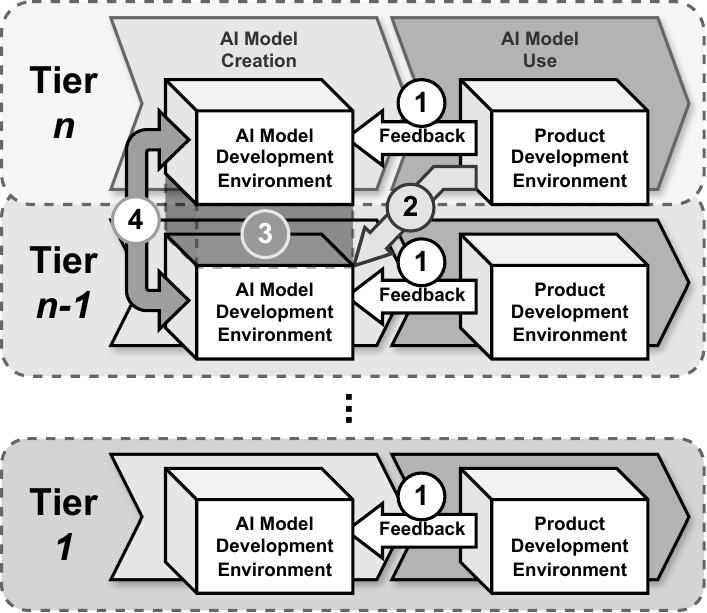}
  \caption{AI model collaboration scenarios}
  \label{figure:usage_scenarios}
\end{figure}

We identified different collaboration scenarios to get a more profound understanding of concepts required to capture all information necessary in collaborative \ac{AI} model development and use. Fig. \ref{figure:usage_scenarios} illustrates the scenarios described in the remainder of this section.

\subsubsection{Internal Use}
\ac{AI} models and datasets remain with the creator.
\ac{AI} models are used from within the \ac{AI} development environment and, for example, within a tier in the automotive industry.
Results are then stored in a product development environment, and feedback is returned to the \ac{AI} development environment, see \circled{1}.

\subsubsection{Shared Data}
If the data cannot be captured internally, an exchange of data between the supply chain participants is an option.
In this scenario, data from the development database can be used in the \ac{AI} development environment, see \circled{2}.
This form of collaboration requires description and well-defined interfaces for data access.

\subsubsection{Shared Models}
Another form of cooperation is sharing trained models.
In contrast to the former cooperation form, no data from the product development environment is shared.
Instead, \ac{AI} models are shared, see \circled{3}.

\subsubsection{Model Services}
\label{subsection:model_services}
The fourth and last cooperation scenario is based on the deployment and utilization of services. Instead of exchanging an \ac{AI} model directly, users send corresponding requests to a service, that delegates to the \ac{AI} model internally, see \circled{4}.

\subsection{Ontology Concepts}
Creating the ontology requires identification of relevant concepts and their relationships. These are then modeled in the ontology as corresponding OWL classes and properties.
Considering the usage scenarios, identified essential concepts are shown in Tab. \ref{tab:ontology_concepts}.
\begin{table}[H]
\centering
\begin{tabular}{@{}ll@{}}
\toprule
\multicolumn{1}{c}{\textbf{Concept}} & \multicolumn{1}{c}{\textbf{Description}}    \\ \midrule
\multicolumn{2}{c}{\cellcolor[HTML]{EFEFEF}\textbf{General metadata}}              \\
Author                               & Provenance information                      \\
Description                          & Natural language description                \\
Name                                 & Name of an AI model or dataset              \\
Framework                            & Back-end used to execute an AI model        \\
\multicolumn{2}{c}{\cellcolor[HTML]{EFEFEF}\textbf{Asset metadata}}                \\
Dataset                              & Data used for training and testing          \\
Data source                          & Location of a dataset                       \\
Training-/test-split                 & Splitting of a dataset                      \\
Data points                          & Size of a dataset                           \\
\multicolumn{2}{c}{\cellcolor[HTML]{EFEFEF}\textbf{AI Model characterization}}     \\
Task / sub-task                      & Intended purpose of an AI model             \\
Input                                & Input of an AI model                        \\
Output                               & Output of an AI model                       \\
Parameter                            & Parameter of an AI model                    \\
Quality criterion                    & Quality criterion of an AI model            \\
Score                                & Score of an AI model regarding a metric    
\end{tabular}
\caption{Essential concepts of the ontology}
\label{tab:ontology_concepts}
\end{table}

%
\section{Evaluation}
\label{section:use_cases}
This section evaluates the proposed approach with the help of a collaboration scenario where typical stakeholders, like domain experts, \ac{AI} experts, and end users, work interdependently and interdisciplinary together. The use case covers the development and usage of an \ac{AI} model that facilitates the prediction of memory access time, which is essential in predicting software timing in safety-critical applications \cite{ottlik2017design}.

The dataset is created by a domain expert or, in our case, a hardware developer. The domain expert describes his inherent knowledge using our proposed ontology \ac{AIMDEO}. In our use case, this covers cache (memory hierarchy) characteristics, such as \emph{replacement strategy} and \emph{size}. The dataset is registered on the \ac{AIMDEP}, easing the \emph{Dataset Exchange}. The \ac{AI} expert can access the dataset and the knowledge specified by the domain expert. This supports the \ac{AI} expert in designing a suitable \ac{AI} model.

The developed model is registered for the \emph{\ac{AI} Model Exchange}, similar to the data set's registration. The \ac{AI} expert provides additional information about the model, which is then annotated according to the ontology for the end user. An end user, in our case a software developer, is looking for an \ac{AI} model to predict software timing. The developer looks for a suitable memory configuration for his or her time-sensitive software for embedded systems. He or she can use the keyword search provided by the platform to find an appropriate model on the platform. The found model can then be used to analyze his software with the memory configuration, as described using the ontology, thus assessing the purchase of test hardware.

The following sections outline this exchange with the mentioned use case.

\subsection{Dataset Exchange}
\label{subsection:dataset_exchange}
The hardware developer publishes the dataset he has created using the proposed platform \ac{AIMDEP}. The platform recognizes all features semi-automatically when the dataset is uploaded, making it easier to describe them. One advantage of automatically recognizing all features of supported data formats is reducing susceptibility to errors, e.g., by forgetting features. In combination with the platform's input mask, the user is also actively supported when entering information, such as a description and the value range. In the subsequent phase, metadata about the dataset and configuration parameters, which in this instance comprise information regarding the memory configuration, such as the \emph{replacement strategy} or the \emph{memory size}, can be specified. This information is then saved using the ontology \ac{AIMDEO} and exported as OWL files upon the download of the dataset. The representation of the dataset in OWL can be observed in Fig. \ref{fig:owl-micro-ontology}, lines 53-61, with further details available in Section \ref{section:ontology}. A search function enables \ac{AI} experts on the platform to locate the dataset easily and quickly. The user does not have to download the dataset manually to analyze it; instead, they can use the analysis functions provided by the platform. This allows for the rapid analysis, comparison, and selection of a suitable dataset, which can then be downloaded together with the description of the dataset by the ontology. This process can also be accomplished via a REST API, enabling easy integration with other tools.

\subsection{\ac{AI} Model Exchange}
\label{subsection:ai_model_exchange}
Once the \ac{AI} expert has developed a suitable \ac{AI} model, it can be published on the platform. Additional information is added to the model using the ontology to enable the end user to quickly and easily assess the model's suitability for the specific application. Firstly, the input and output parameters are described and named. For this purpose, the name, a description, the data type, and optionally, a valid value range and the distribution are specified for each feature; as an example, the information for the \ac{AI} model of the use case can be seen in Fig. \ref{fig:owl-micro-ontology}, lines 4-18. In addition to this information, the uploader can specify configuration parameters according to the scheme in subsection \ref{subsection:dataset_exchange}. With the evaluation information, end users can also assess the quality of the model more efficiently and better, such as using the quality metric and the breakdown of the dataset for training and testing. In addition to this information, meta-information about the model, such as the framework used, a description of the model, and the publication date, is provided. Another essential piece of information is the dataset specification used to assess better the model's suitability for the end users' data.
The end user, in our case, the software developer, can find the model using the platform's built-in search function, which reduces the effort to find an \ac{AI} model. By linking the dataset, the user can also find out what format and quality the training data for the model provided was in and whether their data matches it. The platform provides execution of the model for different frameworks; see Section \ref{section:simml}, which is directly done via the platform. The fast execution makes trying out the model for end users' data possible without setting up a suitable runtime environment. This lowers the hurdle for using and experimenting with an \ac{AI} model. If the model meets the end user's criteria, they can download the model easily together with the ontology description and use it in their workflow.

\begin{figure}
  \subfloat{
\begin{lstlisting}[backgroundcolor=\color{backcolour},language=Turtle,basicstyle=\ttfamily\tiny,numbers=left,numbersep=5pt]
@prefix ns: <http://.../aimodel/> .
@prefix xsd: <http://www.w3.org/2001/XMLSchema#> .

<http://.../aimodel/metric/model0> a ns:AIModel ;
    ns:hasName "RF Instruction Cache-Line Access Classifier" ;
    ns:hasDescription "AI Model to classify single cache-line accesses
                        to the instruction cache as hit or miss." ;
    ns:hasAIModelAuthor <http://.../aimodel/dataset/dataset0/author0> ;
    ns:hasAIModelParameter <http://.../aimodel/parameter/parameter0>,
                            <http://.../aimodel/parameter/parameter1> ;
    ns:hasAIModelInput <http://.../aimodel/input/input0> ;
    ns:hasAIModelOutput <http://.../aimodel/output/output0> ;
    ns:hasAIModelDataset <http://.../aimodel/dataset/dataset0> ;
    ns:hasAIModelMetric <http://.../aimodel/metric/metric0> ;
    ns:hasFramework "scikit-learn" ;
    ns:hasTask "https://identifiers.org/ito:ITO_42033" ;
    ns:hasSubtask "https://identifiers.org/ito:ITO_00498" ;
    ns:hasVersion "1.0" .

<http://.../aimodel/metric/metric0> a ns:AIModelMetric ;
    ns:hasName "average precision" ;
    ns:hasDescription "Model evaluation metric" ;
    ns:hasScore "0.9794" ;
    ns:hasUnit "dimensionless" .

<http://.../aimodel/parameter/parameter0> a ns:AIModelParameter ;
    ns:hasName "Replacement-Strategy" ;
    ns:hasDescription "Replacement-Strategy of the used Instruction Cache" ;
    ns:hasUnit "string" ;
    ns:hasScore "LRU" .

<http://.../aimodel/parameter/parameter1> a ns:AIModelParameter ;
    ns:hasName "Cache-Size" ;
    ns:hasDescription "Size of the used Instruction Cache" ;
    ns:hasUnit "Byte" ;
    ns:hasScore "2048" .

<http://.../aimodel/input/input0> a ns:AIModelInput ;
    ns:hasName "set" ;
    ns:hasDescription "Set of the cache-line" ;
    ns:hasUnit "uint8" .

<http://.../aimodel/output/output0> a ns:AIModelOutput ;
    ns:hasName "miss" ;
    ns:hasDescription "Category, if the access to the cache-line was
                        a hit (false) or a miss (true)" ;
    ns:hasUnit "bool" .

<http://.../aimodel/dataset/dataset0/source0> a ns:AIModelDatasetSource ;
    ns:hasName "Cache-accesses database location" .
    ns:hasLocation "file://.../datasets/0/hbu9vRD.csv" ;

<http://.../aimodel/dataset/dataset0> a ns:AIModelDataset ;
    ns:hasAIModelDatasetAuthor <http://.../aimodel/dataset/dataset0/author1> ;
    ns:hasAIModelDatasetSource <http://.../aimodel/dataset/dataset0/source0> ;
    ns:hasDescription "Contains records of instruction-cache-accesses
                        for different generated complex pseudo-programs." ;
    ns:hasName "Complex Programs Instruction Cache-Lines Set Sorted" ;
    ns:hasTask "https://identifiers.org/ito:ITO_42033" ;
    ns:hasSubtask "https://identifiers.org/ito:ITO_00498" ;
    ns:hasVersion "3.0" .
\end{lstlisting}}
  \caption{Micro-ontology describing \ac{AI} model metadata}
  \label{fig:owl-micro-ontology}
\end{figure}

%
\section{Conclusion}
\label{section:conclusion}
We used the \ac{AIMDEP} successfully to collaboratively develop \ac{AI} models. However, its application is not limited to that. It is possible to use either the platform or the ontology separately.
The platform eases collaborative \ac{AI} model development while the ontology can help to ensure that required information is available, and how this information can be provided.
The metadata specification and export functionality of the platform can help in integrating different tools required in the \ac{AI} model development process.

Future work could, for instance, investigate on extending \ac{EMMM} to create a certain degree of compatibility.
The generic key-value pair metadata associated with the resource types in the metamodel could be used to store \ac{AIMDEO} metadata.
Creating further interoperability between \ac{AIMDEO} and \ac{ITO} might also be worth investigating.

\balance{}

\section*{Acknowledgment}
This paper is funded by the BMWi within the project progressivKI (grant number 19A21006M).

\bibliographystyle{IEEEtran}
\bibliography{main}

@article{blagec2022curated,
  title={A curated, ontology-based, large-scale knowledge graph of artificial intelligence tasks and benchmarks},
  author={Blagec, Kathrin and Barbosa-Silva, Adriano and Ott, Simon and Samwald, Matthias},
  journal={Scientific Data},
  volume={9},
  number={1},
  pages={322},
  year={2022},
  publisher={Nature Publishing Group UK London}
}

@article{idowu2022asset,
  title={Asset management in machine learning: State-of-research and state-of-practice},
  author={Idowu, Samuel and Strüber, Daniel and Berger, Thorsten},
  journal={ACM Computing Surveys},
  volume={55},
  number={7},
  pages={1--35},
  year={2022},
  publisher={ACM New York, NY}
}

@inproceedings{idowu2022emmm,
  title={{EMMM}: A unified meta-model for tracking machine learning experiments},
  author={Idowu, Samuel and Str{\"u}ber, Daniel and Berger, Thorsten},
  booktitle={2022 48th Euromicro Conference on Software Engineering and Advanced Applications (SEAA)},
  pages={48--55},
  year={2022},
  organization={IEEE}
}

@online{mlem, author = {iterative.ai}, title = {MLEM}, url = {https://mlem.ai} }

@INPROCEEDINGS{ottlik2017design,
  author={Ottlik, Sebastian and Gerum, Christoph and Viehl, Alexander and Rosenstiel, Wolfgang and Bringmann, Oliver},
  booktitle={Design, Automation \& Test in Europe Conference \& Exhibition (DATE), 2017}, 
  title={Context-sensitive timing automata for fast source level simulation}, 
  year={2017},
  volume={},
  number={},
  pages={512-517},
  doi={10.23919/DATE.2017.7927042}}

@online{plotly, author = {Plotly Technologies Inc.}, title = {Collaborative data science}, publisher = {Plotly Technologies Inc.}, address = {Montreal, QC}, year = {2015}, url = {https://plot.ly} }

@incollection{antoniou2004webOntologyLanguage,
  title={Web ontology language: Owl},
  author={Antoniou, Grigoris and Van Harmelen, Frank},
  booktitle={Handbook on ontologies},
  pages={67--92},
  year={2004},
  publisher={Springer}
}

\end{document}